%% file: main.tex
\documentclass[10pt,twocolumn,letterpaper]{article}

\usepackage[table]{xcolor}
\usepackage[pagenumbers]{wacv}
\usepackage{times}
\usepackage{epsfig}
\usepackage{graphicx}
\usepackage{amsmath}
\usepackage{amssymb}
\usepackage{booktabs}
\usepackage{custom_commands}
\usepackage{float}
\usepackage{tabularray}
\usepackage{makecell}
\usepackage{algorithm}
\usepackage{algpseudocode}
\usepackage[nolist]{acronym}
\usepackage{pifont}
\usepackage{tikz}
\usepackage{svg}
\newcommand{\cmark}{\ding{51}}%

\makeatletter
\newcommand*{\org@overidelabel}{}
\let\org@overridelabel\@verridelabel
\@ifpackagelater{acronym}{2015/03/21}{
  \renewcommand*{\@verridelabel}[1]{%
    \@bsphack
    \protected@write\@auxout{}{\string\AC@undonewlabel{#1@cref}}%
    \org@overridelabel{#1}%
    \@esphack
  }%
}{
  \renewcommand*{\@verridelabel}[1]{%
    \@bsphack
    \protected@write\@auxout{}{\string\undonewlabel{#1@cref}}%
    \org@overridelabel{#1}%
    \@esphack
  }%
}
\makeatother

\usetikzlibrary{decorations.pathreplacing, positioning, arrows.meta, calc}

\usepackage{listings}
\lstdefinestyle{mystyle}{
    basicstyle=\ttfamily\footnotesize,
    breakatwhitespace=false,         
    breaklines=true,                 
    captionpos=b,                    
    numbersep=5pt,                  
    showspaces=false,                
    showstringspaces=false,
    showtabs=false,                  
    tabsize=2
}

\usepackage[pagebackref=true,breaklinks=true,colorlinks,bookmarks=false]{hyperref}
\hypersetup{
    colorlinks = true,
    citecolor = {black},
    linkcolor = {black},
    urlcolor = {black},
}

\newcommand\graycite[1]{\hypersetup{citecolor=black!40}\cite{#1}\hypersetup{citecolor=black}}

\usepackage[capitalize]{cleveref}
\usepackage{orcidlink}
\crefname{section}{Sec.}{Secs.}
\Crefname{section}{Section}{Sections}
\Crefname{table}{Table}{Tables}
\crefname{table}{Tab.}{Tabs.}

\def\wacvPaperID{1698} 
\def\confName{WACV}
\def\confYear{2024}

\begin{document}

\setlength{\abovedisplayskip}{6pt}
\setlength{\belowdisplayskip}{6pt}

\input{tex/99-acronyms}

\title{BASE: Probably a Better Approach to Multi-Object Tracking}

\author{
Martin Vonheim Larsen%
\thanks{
This work was funded by The University Center at Kjeller, and by projects 1505 and 1688 at the Norwegian Defence Research Establishment.
}
 $\, ^{1,2}$,
Sigmund Rolfsjord$^{1,2}$,
Daniel Gusland$^{1,2}$,\\
Jörgen Ahlberg$^{3,2}$
and Kim Mathiassen$^{1,2}$
\\
\\
Norwegian Defence Research Establishment$^1$,
University of Oslo$^2$,
Linköping University$^3$
\\
{\tt\small\href{mailto:martin-vonheim.larsen@ffi.no}{martin-vonheim.larsen@ffi.no}}
}%
\maketitle

\begin{abstract}
The field of visual object tracking is dominated by methods that combine simple tracking algorithms and ad hoc schemes.
Probabilistic tracking algorithms, which are leading in other fields, are surprisingly absent from the leaderboards. 
We found that accounting for distance in target kinematics, exploiting detector confidence and modelling non-uniform clutter characteristics is critical for a probabilistic tracker to work in visual tracking.
Previous probabilistic methods fail to address most or all these aspects, which we believe is why they fall so far behind current state-of-the-art (SOTA) methods (there are no probabilistic trackers in the MOT17 top 100).
To rekindle progress among probabilistic approaches, we propose a set of pragmatic models addressing these challenges, and demonstrate how they can be incorporated into a probabilistic framework.
We present \acs{BASE} (\acl{BASE}), a simple, performant and easily extendible visual tracker, achieving \ac{SOTA} on MOT17 and MOT20, without using Re-Id.
Code will be made available at \url{https://github.com/ffi-no}.
\end{abstract}

\acresetall

\input{tex/01-introduction}

\input{tex/02-related-work}
\input{tex/03-background}
\input{tex/04-implementation}
\input{tex/05-experiments}
\input{tex/06-conclusion}
{\small
\bibliographystyle{utils/ieee_fullname}

\input{main.bbl}
}

\end{document}

%% file: tex/99-acronyms.tex
\begin{acronym}
\acro{BASE}{Bayesian Approximation Single-hypothesis Estimator}
\acro{PDA}{Probabilistic Data Association}
\acro{JPDA}{Joint Probabilistic Data Association}
\acro{GNN}{Global Nearest Neighbor}
\acro{IF}{information filter}
\acro{IOU}{intersection-over-union}
\acro{KF}{Kalman filter}
\acro{LR}{likelihood-ratio}
\acro{MHT}{Multi-Hypothesis Tracking}
\acro{MLE}{maximum likelihood estimation}
\acro{NCV}{nearly constant velocity}
\acro{PHD}{probability hypothesis density}
\acro{RFS}{Random Finite Sets}
\acro{SHT}{single-hypothesis tracker}
\acro{SORT}{Simple Online and Realtime Tracking}
\acro{SOTA}{state-of-the-art}
\acro{VMOT}{Visual Multi-Object Tracking}
\acro{CMC}{camera motion compensation}
\acro{FN}{false negatives}
\acro{VO}{visual odometry}
\end{acronym}

%% file: tex/01-introduction.tex
\section{Introduction}
\label{sec:introduction}
\begin{figure}
    \centering
    \includegraphics[width=1.0\linewidth]{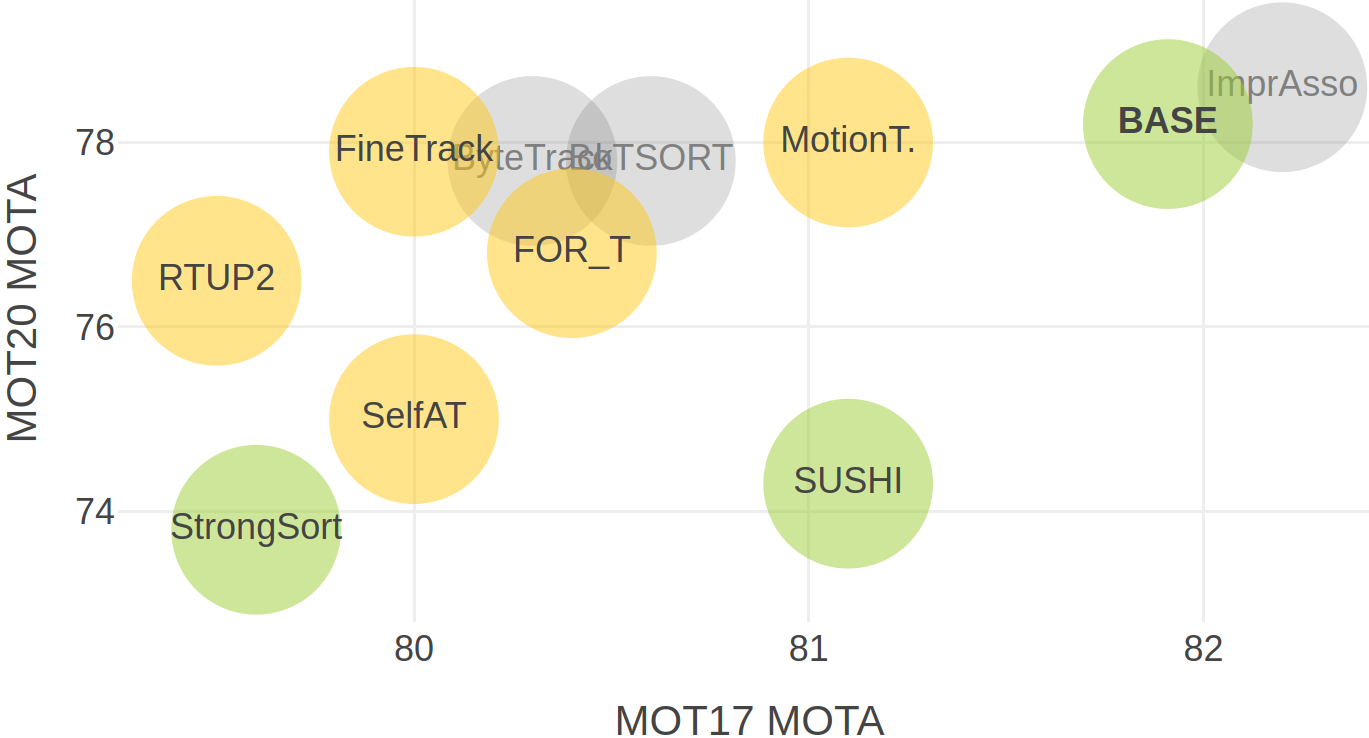}
    \caption{\label{fig:mot-results}
    MOTA comparison of our proposed method BASE and top-performing trackers on the MOT17 and MOT20 benchmarks.
    Trackers in green use a fixed set of parameters across the test set, while those in grey tune parameters for each test sequence.
    Yellow trackers do not report whether parameters are kept constant on the test set, and have not published code reproducing the results.
    }
\end{figure}

\Ac{VMOT} is the task of estimating the location of objects over time in a video sequence while maintaining a unique ID for each target. 
Popular \Ac{VMOT} benchmarks \cite{Leal-Taixe2015MOTChallengeTracking,Dendorfer2020MOT20:Scenes,Sun2022DanceTrack:Motion} are currently dominated by methods which combine simple tracking algorithms with a stack of ad hoc specializations to visual tracking \cite{Zhang2021ByteTrack:Box, Du2022StrongSORT:Again,Aharon2022BoT-SORT:Tracking,Yang2023HardSpace}.
These simple-yet-effective trackers cut corners using hard logic, for instance by ignoring less-certain detections, leaving performance on the table.
Meanwhile, \textit{probabilistic trackers} are ubiquitous in more mature fields such as radar- and sonar tracking, as they avoid most of these hard choices and can better exploit the available information.
This raises the question: \textit{Why are probabilistic methods outperformed by ad hoc approaches on \ac{VMOT}?}
We believe the probabilistic approaches have overlooked a few key aspects specific to \ac{VMOT} in their adaptations of existing tracking theory.

In visual tracking, the perspective imaging results in target kinematics and clutter (false detections) characteristics that are radically different from those seen in radar or sonar tracking.
When representing target kinematics in image plane coordinates, we should expect objects near the camera to appear more agile than those we see from afar.
Similarly, we should expect the density of new targets and clutter detections to be much greater for distant objects, based on the simple fact that objects take up less space in the image when they are farther away.
Accounting for these non-uniform effects is necessary to succeed in probabilistic visual tracking.

The dominating ad hoc trackers are pragmatically built from the ground up for \ac{VMOT}.
One example of this is the ubiquitous use of \ac{IOU} as a similarity metric for detection-to-track matching.
Although chosen for its simplicity, the \ac{IOU} metric has a nice side-effect in that it gives more leeway for larger bounding boxes than for smaller ones.
Leading approaches \cite{Aharon2022BoT-SORT:Tracking,Yang2023HardSpace} also use object size to change model dynamics when estimating the motion of targets.
Avoiding the drawbacks of modeling distance explicitly, these pragmatic mechanisms lead to target models that are \textit{distance-aware}, with good performance both for near and distant objects.


The advantage probabilistic trackers hold over ad hoc methods lies in their ability to better balance different types of information.
Which detection is likely to originate from the current object? 
The one closest in position, the one with the highest confidence, or perhaps the one most similar in size or appearance?
Ad hoc methods either ignore parts of this information, or resort to some arbitrary weighting between them.
Similarly, ad hoc methods typically accept targets once they have been detected a fixed number of times, ignoring how consistent the detections were, or what confidence level they had.
A strong probabilistic approach would instead model the relevant aspects of this information, and take decisions based on what is most \textit{likely}.

\begin{figure*}[ht]
\vspace{-12pt}
    \centering
    \begin{tikzpicture}
    \begin{scope}
    \clip [rounded corners=.2cm] (0,0) rectangle coordinate (centerpoint) (\linewidth, 170pt); 
    \node [inner sep=0pt] at ($(centerpoint) + (0.4,-1.95)$) {
    \includesvg[width=1.05\linewidth]{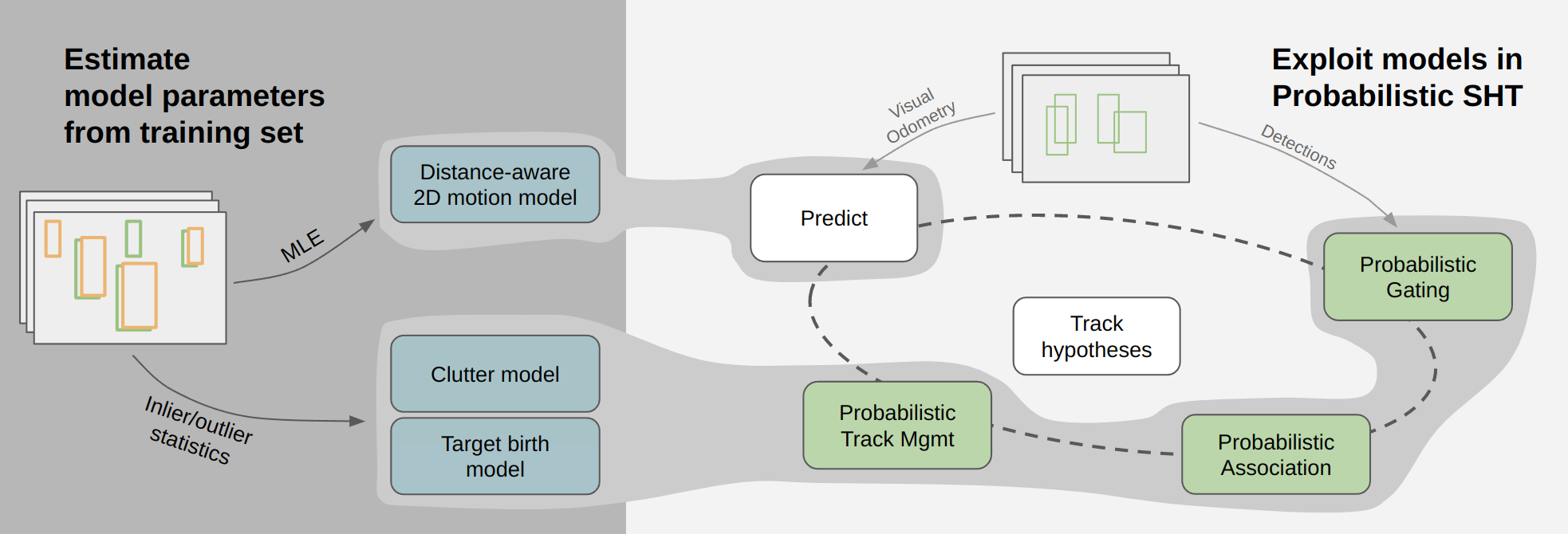}
    }; 
    \end{scope}
    \end{tikzpicture}
    \caption{
    \ac{BASE} builds on a traditional \acf{SHT} architecture, but uses probabilistic formulations for all aspects of gating, association and track management, as well as a \textit{distance-aware} motion model.
    These probabilistic formulations enable \ac{BASE} to express nuances in detection confidence and detection-to-track match, which threshold-based approaches lack.
    For a given usecase, \ac{BASE} requires modeling the motion, clutter and target birth characteristics of the camera and detection algorithm used, which we can automatically fit using typical training datasets.
    }
    \label{fig:base-overview}
\end{figure*}

Despite the more complex structure of probabilistic trackers, we find that the leading approaches \cite{Fu2019Multi-LevelTracking, Song2019OnlineManagement, JinlongYang2022ImprovingFilters, Baisa2021RobustLearning} omit, or fail, to model the aspects we see as critical to visual tracking.
To lift probabilistic trackers to the performance of the leading ad hoc methods, we propose \textit{\ac{BASE}}, a minimalist probabilistic take on visual tracking.
As outlined in \cref{fig:base-overview}, \ac{BASE} replaces the key components of a traditional \ac{SHT} with probabilistic counterparts.
The novelty of our approach is accounting for the non-uniform kinematics and distribution of clutter experienced in visual tracking, in a unified probabilistic manner.
Specifically, our main contributions are:
\begin{itemize}
    \setlength\itemsep{-0.3em}
    \item A distance-aware motion model.
    \item Pragmatic modelling of new targets and clutter detections, suited for \ac{VMOT}.
    \item A new \textit{association-less} probabilistic track management scheme, suited for the crowded scenes in \ac{VMOT}.
    \item Methods for automatically estimating model parameters from training data.
\end{itemize}

\Cref{sec:related-work} gives a brief review of related work, followed by \cref{sec:sht-revisit} which revisits the traditional single-hypothesis tracking pipeline. 
\Cref{sec:method} describes our proposed method, BASE, and \cref{sec:experiments} details our experiments performed on MOT17, MOT20 and DanceTrack.  
\Cref{sec:conclusione} contains a summary and conclusions.

%% file: tex/02-related-work.tex
\section{Related work}
\label{sec:related-work}
\noindent\textbf{Visual multi-object tracking.}
The simple and performant trackers that are currently \ac{SOTA} are mostly based on or inspired by \ac{SORT} \cite{Bewley2016SimpleTracking}. 
SORT \cite{Bewley2016SimpleTracking} started a strong line of pragmatic trackers \cite{Wojke2018SimpleMetric,Du2022StrongSORT:Again,Zhang2021ByteTrack:Box, Aharon2022BoT-SORT:Tracking,Cao2022Observation-CentricTracking,Nasseri2022FastTracking,Yang2023HardSpace, Stadler2022ModellingCrowds}.
These publications have a strong empirical focus, identifying more or less stand-alone components that improve performance.
Based on their research, a set of core components have emerged that are common among strong visual trackers.
Some of these components are \ac{GNN} matching based on IOU-related association metrics, Kalman filtering for track prediction, two-stage track formation, ego-motion compensation, post-interpolation of missed segments, and most importantly, a strong detector. 

\noindent\textbf{End-to-end-trackers} trackers are very attractive, as they can be adapted to new applications without architectural changes.
They can also take advantage of information in ways that are hard to express in code.
Although many end-to-end trackers exhibit impressive results \cite{Sun2020TransTrack:Transformer, Zhu2021LookingTransformers,Zeng, Yang2021ReMOT:Tracking,Sun2020TransTrack:Transformer,Yan2022TowardsTracking,Wang2020JointNetworks, Zhang2023MOTRv2:Detectors}, they lag behind the best tracking-by-detection methods on many \ac{VMOT} benchmarks like MOTChallenge.
This is perhaps due to limited training data and difficulties in training motion models with a small time horizon \cite{Zhu2021LookingTransformers}. 
These issues may be resolved in the future, but we still believe there is room for other approaches, as data restrictions will likely persist in many niche applications.

\noindent\textbf{Probabilistic visual tracking.}
Many probabilistic trackers in the MOTChallenges build on the ideas of Bar-Shalom \cite{Bar-Shalom1975TrackingAssociation,Bar-Shalom2007DimensionlessTracking} and Blackman 
 \cite{blackman1999design}.
While they mostly rely on advanced state management schemes like MHT \cite{Reid1979AnTargets,Kim2015MultipleRevisited} or PHD \cite{Mahler2003MultitargetMoments}, they often resort to non-probabilistic methods for association and for combining different types of information, such as appearance or shape.
In fact, the most successful probabilistic attempts fall back to non-probabilistic association methods like IOU matching.
Meanwhile, approaches which rely solely on probabilistic schemes \cite{Baisa2021RobustLearning}, score slightly worse than even the extremely simple IOUTracker \cite{Bochinski2017High-SpeedInformation}.

Surprisingly, most probabilistic approaches do not leverage detector confidence beyond basic detection thresholding \cite{Sanchez-Matilla2016OnlineDetections, Fu2019Multi-LevelTracking,JinlongYang2022ImprovingFilters,Fu2018ParticleLearning,Aguilar2022SmallFilter,Baisa2021RobustLearning}.
Some approaches utilize confidence score for track initiation only \cite{Song2019OnlineManagement, Baisa2021Occlusion-robustRe-identification,Baisa2019OnlineLearning}.
Meanwhile, Wojke et al. \cite{Wojke2017Confidence-AwareTracking} demonstrated a significant boost in performance by integrating detector confidence in a PHD-filter.

%% file: tex/03-background.tex
\section{Probabilistic SHT tracking revisited}
\label{sec:sht-revisit}
To bridge the gap between the highly specialized ad hoc methods and the overly general probabilistic trackers, the \acf{SHT} is a good starting point.
It is arguably one of the simplest tracking methods that can also be made to leverage most of the key building blocks of probabilistic trackers.

Traditional \ac{SHT} can be summarized as developing a single set of track hypotheses through the following steps for each new piece of sensor data:

\begin{enumerate}
    \setlength\itemsep{-0.3em}
    \item \textbf{Predict} existing tracks to the current timestep.
    \item \textbf{Gate} the detections by disregarding detection-to-track pairs that are considered too unlikely to be associated.
    \item \textbf{Associate} the overall most likely connections between detections and existing track hypothesis.
    \item \textbf{Update} existing tracks using associated detection data.
    \item \textbf{Manage tracks} by initializing new tracks for unmatched detections and discarding unlikely tracks.
\end{enumerate}
This structure is popular among the top-performing visual trackers, which implement most (or all) these steps using ad hoc modelling \cite{Zhang2021ByteTrack:Box,Aharon2022BoT-SORT:Tracking,Du2022StrongSORT:Again}.
\textit{Probabilistic \ac{SHT}} aims to formulate gating, association and track management in terms of probability.
These formulations often also require strict probabilistic modeling of the underlying state (predict and update).
Blackman \cite{blackman1999design} gives an excellent in-depth walkthrough of probabilistic \ac{SHT}, but we will here go through the most critical aspects.

\subsection{Probabilistic gating and association}
The ad hoc trackers discussed in \cref{sec:related-work} typically perform gating by thresholding the detection-to-track IOU distance, and association by minimizing the overall IOU distance.
The benefit of formulating these mechanisms in terms of probability is that it offers a clear path to incorporate aspects such as the quality of the track estimates, the detector performance, and the clutter characteristics.

For a given true track state $\vecx_i$, we assume to have an estimate $\widehat{\vecx}_i \sim \mathcal{N}\left( \vecx_i,\, \matP_i\right)$ and that corresponding detections $\vecz_j$ are generated as
\begin{align}
\vecz_j = \matH\vecx_i + \vecw_j,\, \vecw_j \sim \mathcal{N}\left( \veczero, \matR_j \right). \label{eq:z-obs}
\end{align}
Here, $\matH$ is the measurement function, $\vecw_j$ is white measurement noise, and $\matP_i$ and $\matR_j$ are known covariances.
We can then consider the detection-to-track innovation
\begin{align}
\widehat{\vecy}_{ij} = \vecz_j - \matH\widehat{\vecx}_i \sim \mathcal{N}\left( \veczero, \matS_{ij}\right), \label{eq:y-inno}
\end{align}
with $\matS_{ij} = \matR_j + \matH \matP_i \matH\trans$.

For gating, an traditional approach (see \cite[Sec. 6.3.2]{blackman1999design} for details) is to threshold the match-to-noise likelihood ratio:
\begin{align}
\frac{\mathcal{N}\left(\widehat{\vecy}_{ij};\, \veczero, \matS_{ij}\right)}{\lambda_C + \lambda_{NT}} \geq \frac{1 - P_G}{P_G}, \label{eq:sht-gate}
\end{align}
where $\lambda_{NT}$ and $\lambda_C$ are the densities of new targets and clutter, and $P_G$ is the desired gating confidence level.

For probabilistic association, we want to find the ``most likely'' set of detection-to-track pairs $\mathcal{A} = \{(j, i)\}$.
A common approach, which ignores the detector performance and clutter characteristics, is to formulate this as linear-sum assignment problem over the negative logarithm of the innovation likelihood:
\begin{align}
\mathcal{A}^* = \argmin_\mathcal{A} \sum\limits_{(j, i) \in \mathcal{A}} \widehat{\vecy}_{ij}\trans \matS_{ij}^{-1} \widehat{\vecy}_{ij} + \log |\matS_{ij}| \label{eq:sht-assoc}
\end{align}

\subsection{Probabilistic track management}
Where ad-hoc trackers employ schemes such as counting the number of recent detections to assess whether a track hypothesis should be discarded, probabilistic \ac{SHT} estimates the probability that each hypothesis is valid.
This is done by assessing the event
\begin{equation*}
\begin{aligned}
\mathcal{H}_i:\quad x_i\, &\text{was either not detected, or explained by a}\\
&\text{detection from a real target, in each frame}.
\end{aligned}
\end{equation*}
For each track $i$ we then maintain a \ac{LR} $\mathrm{LR}_i$, weighing evidence for- and against $\mathcal{H}_i$, as
\begin{align}
\mathrm{LR}_i \triangleq \frac{\clik{\mathcal{Z}_k}{\mathcal{H}_i}}{\clik{\mathcal{Z}_k}{\overline{\mathcal{H}}_i}} \cdots \frac{\clik{\mathcal{Z}_1}{\mathcal{H}_i}}{\clik{\mathcal{Z}_1}{\overline{\mathcal{H}}_i}}\frac{\prob{\mathcal{H}_i}}{\prob{\overline{\mathcal{H}}_i}}, \label{eq:hyp-test}
\end{align}
where $\mathcal{Z}_k$ represents the sensor data at time $k$, and $\overline{\mathcal{H}}_i$ is the logical complement of $\mathcal{H}_i$.
At each timestep $k$, depending on whether the track was detected or not, Blackman writes the corresponding \ac{LR} factor as
\begin{equation}
\begin{aligned}
\frac{\clik{\mathcal{Z}_k}{\mathcal{H}_i}}{\clik{\mathcal{Z}_k}{\overline{\mathcal{H}_i}}}
&=
\begin{cases}
    P_D\frac{\lik{\widehat{\vecy}_{ij}}}{\lambda_C}
\frac{\clik{z_S}{\mathrm{Det}, \mathcal{H}}}{\clik{z_S}{\mathrm{Det}, \overline{\mathcal{H}}}}, & \text{if assoc. to }j\\
    1 - P_D,& \text{otherwise.}
\end{cases}
\end{aligned}
\label{eq:blackman-llr}
\end{equation}
Here, $P_D$ is prior detection probability, which is typically modelled as a constant.
$\mathrm{LR}_S = \frac{\clik{z_S}{\mathrm{Det}, \mathcal{H}}}{\clik{z_S}{\mathrm{Det}, \overline{\mathcal{H}}}}$ is the ``signal-related'' \ac{LR}, typically derived from the SNR of the given detection method.
We then consider a track hypothesis $x_i$ to be ``unconfirmed'' until $\mathrm{LR}_i$ passes some given threshold, and discard the hypothesis if $\mathrm{LR}_i$ falls below some other threshold.



%% file: tex/04-implementation.tex
\section{The BASE visual tracker}\label{sec:method}
In this section, we present \ac{BASE}, a minimalist probabilistic take on visual tracking.
We design \ac{BASE} as a probabilistic \ac{SHT} with the necessary extensions to sufficiently model the visual tracking problem, as shown in \Cref{fig:base-overview}.

To accommodate the non-uniform motion and clutter encountered in visual tracking, we develop the \textit{distance-aware 2D motion model} in \cref{sec:motion-sensor-model} and model detector performance and clutter behaviour in \cref{sec:clutter-model}.
In \cref{sec:estimate-motion} we propose an automatic procedure to estimate the parameters of these models.
The traditional \ac{SHT} pipeline discussed in \cref{sec:sht-revisit} cannot fully exploit these models in all aspects of gating, association and track management.
We therefore compute the \textit{association probability} in \cref{sec:assoc-prob}, before using it to reformulate gating and association in \cref{sec:gate-and-assoc} and track management in \cref{sec:track-mgmt}.

\subsection{Estimating the association probability}\label{sec:assoc-prob}
To account for non-uniform densities of new targets and clutter, the traditional association score from \cref{eq:sht-assoc} is insufficient.
Consider a track which is presented with two measurements that have identical statistical distance, but where one measurement has far greater risk of being clutter than the other.
Intuitively, we should prefer to associate the track with the measurement least likely to be clutter.

Instead of using the traditional association score from \cref{eq:sht-assoc}, we shall compute the full \textit{association probability} for each detection/track pair.
First, we define the event
\begin{equation*}
\mathcal{A}_{ij}:\quad\, z_j \text{ originates from the real target represented by }x_i.
\end{equation*}
In our single-hypothesis paradigm, each detection $z_j$ must originate either from a target for which we have an hypothesis, a newly appeared target, or from clutter.
Given the corresponding measurement $\vecz_j$ we use 
\begin{align}
\lambda_\mathrm{EX}\left(\vecz_j\right) = \lambda_\mathrm{NT}\left(\vecz_j\right) + \lambda_\mathrm{C}\left(\vecz_j\right) \label{eq:lambda-ex}
\end{align}
to denote the corresponding \textit{extraneous measurement} density.
Modelling $\lambda_\mathrm{NT}$ and $\lambda_\mathrm{C}$ is detailed in \cref{sec:clutter-model}.

We can now write the association probability as
\begin{align}
\cprob{\mathcal{A}_{ij}}{\widehat{\vecx}_i, \vecz_j}
&= \frac{\clik{\widehat{\vecx}_i, \vecz_j}{\mathcal{A}_{ij}}}{\lambda_\mathrm{EX}\left(\vecz_j\right) + \sum_l \clik{\widehat{\vecx}_l, \vecz_j}{\mathcal{A}_{ij}}}.\label{eq:assoc-lik}
\end{align}
Here, $\clik{\widehat{\vecx}_i, \vecz_j}{\mathcal{A}_{ij}}$ is the joint likelihood of the current track state $\widehat{\vecx}_i$ and the observed measurement $\vecz_j$, assuming that $z_j$ originates from $x_i$.
Inspired by \cite{blackman1999design} we model our measurements to consist of a \textit{state-related measurement} (modelled in \cref{sec:motion-sensor-model}) and an independent \textit{confidence measurement} (modelled in \cref{sec:clutter-model}).
Since we will only model the bounding box state of each track, we write
\begin{align}
\clik{\widehat{\vecx}_i, \vecz_j}{\mathcal{A}_{ij}} \triangleq
\clik{\widehat{\vecx}^{(\mathrm{bb})}_i, \vecz^{(\mathrm{bb})}_j}{\mathcal{A}_{ij}} \lik{\vecz^{(\mathrm{c})}_j}. \label{eq:det-track-lik}
\end{align}


\subsection{Probabilistic gating and association}\label{sec:gate-and-assoc}
Instead of using the traditional \cref{eq:sht-gate} for gating and \cref{eq:sht-assoc} for association scores, we will base both gating and scoring on \cref{eq:assoc-lik}.
We compute association scores as
\begin{align}
- \log \cprob{\mathcal{A}_{ij}}{\widehat{\vecx}_i, \vecz_j} \label{eq:assoc-score}
\end{align}
and perform gating using
\begin{align}
\cprob{\mathcal{A}_{ij}}{\widehat{\vecx}_i, \vecz_j} \geq \frac{1 - P_G}{P_G}.
\end{align}


The introduction of \cref{eq:assoc-lik} results in stricter than before gates both for measurements in crowded regions and less confident measurements.
For association, however, \cref{eq:assoc-lik} also takes the extraneous measurement density into account, critical to properly balance between measurements that have vastly different $\lambda_\mathrm{EX}$.
Using \cref{eq:assoc-score}, the linear-sum assignment will now find the set of associations with the overall lowest probability of containing a misassociation, whereas the traditional variant (\cref{eq:sht-assoc}) finds the most likely association only with regards to the predicted state vs the observed measurements.

\subsection{Probabilistic track management}\label{sec:track-mgmt}
In \ac{BASE}, we will build the track management around the probability that \textit{at least one of the measurements originates from a given track $x_i$} in the current frame, namely
\begin{align}
\widetilde{\mathrm{P}}_i &= 1 - \prod_j \left( 1 - \cprob{\mathcal{A}_{ij}}{\widehat{\vecx}_i, \vecz_j}\right). \label{eq:ptilde}
\end{align}
To rewrite \cref{eq:hyp-test} using $\widetilde{\mathrm{P}}_i$, we first define the event
\begin{equation}
\begin{aligned}
\mathcal{D}^{(k)}_i:\quad x_i\, &\text{was detected in frame }k.
\end{aligned}
\end{equation}
Re-ordering \cref{eq:hyp-test} with Bayes' rule, we can write the \ac{LR}-contribution of frame $k$ as
\begin{align}
\medmuskip=2mu   
\thickmuskip=3mu 
\renewcommand\arraystretch{1.5}
\frac{\clik{\mathcal{H}_i}{\mathcal{Z}_k}}{\clik{\overline{\mathcal{H}_i}}{\mathcal{Z}_k}} 
&= \frac{\clik{\mathcal{H}_i, \mathcal{D}^{(k)}_i}{\mathcal{Z}_k} + \clik{\mathcal{H}_i, \overline{\mathcal{D}}^{(k)}_i}{\mathcal{Z}_k}}{\clik{\overline{\mathcal{H}}_i, \mathcal{D}^{(k)}_i}{\mathcal{Z}_k} + \clik{\overline{\mathcal{H}}_i, \overline{\mathcal{D}}^{(k)}_i}{\mathcal{Z}_k}}\\
&= \frac{\widetilde{\mathrm{P}}_i^{(k)} + \left(1 - P_\mathrm{D}\right)\left(1 - \widetilde{\mathrm{P}}_i^{(k)}\right)}{P_D \left(1 - \widetilde{\mathrm{P}}_i^{(k)}\right)}, \label{eq:delta-llr}
\end{align}
where we have used that $\mathcal{D}^{(k)}_i \subset \mathcal{H}_i$, and that
\begin{align}
P_D \triangleq \cprob{\mathcal{D}^{(k)}_i}{\mathcal{H}_i}.
\end{align}

Where the traditional \ac{LR} update (\cref{eq:blackman-llr}) only uses the associated measurement, \cref{eq:delta-llr} collects contributions from \textit{all} measurements, normalized across all tracks.
This \textit{association-less} track \ac{LR} better handles the cases where several tracks have significant stakes in a given measurement, which is often the case in crowded visual tracking.
Since this computation is well-suited for GPU acceleration and can be run in parallel with the association, the increased computational burden is more than made up for in practice.

\subsection{The distance-aware planar motion model}\label{sec:motion-sensor-model}
%
We shall stick to a traditional linear \ac{KF} setup to model measurements and target motion.
For each target, we model the bounding box center, width and height in the current image plane using pixel coordinates.
We assume a \ac{NCV} model for the bounding box center $c_x, c_y$, and use a nearly constant state model for the box size.
As a state vector we choose
\begin{align}
\vecx =
\begin{pmatrix}
c_x,
\, c_y,
\, \dot{c}_x,
\, \dot{c}_y,
\, w,
\, h
\end{pmatrix},
\end{align}
where $\dot{c}_x, \dot{c}_y$ denote the center velocity, and $w$ and $h$ is the width and height.
We model each \ac{NCV} block using
\begin{align}
\matF_{cv} = \begin{bmatrix}
1 & \delta_t \\
0 & 1
\end{bmatrix},
\matQ_{cv} = \begin{bmatrix}
\delta_t^3/3 & \delta_t^2/2 \\
 \delta_t^2/2 & \delta_t
\end{bmatrix},
\end{align}
for a given timestep $\delta_t$, as in \cite[Sec. 4.2.2]{blackman1999design}.
We use $\matF_{cv}^{*2}$ and $\matQ_{cv}^{*2}$ to denote the 2D composition of $\matF_{cv}$ and $\matQ_{cv}$.

We correct for camera ego-motion using the same method as \cite{Aharon2022BoT-SORT:Tracking}.
Between each pair of following images we obtain a transform which predicts pixels in time as
\begin{align}
\vecp_k = \matW_k \vecp_{k - 1} + \vect_k.
\end{align}
By using the notation $\matT_k = \diag{\matW_k, \matW_k, \matW_k}$ and $\matF_k = \diag{\matF_{cv}^{*2}, \matI_2}$,
we can write the state transition as
\begin{align}
\vecx_{k} &= \matT_k(\matF_k\vecx_{k-1} + \vecv_k) + \matI_{6\times2}\vect_k,
\end{align}
with white $\vecv_k \sim \mathcal{N}(\veczero, \matQ_k)$, where
\begin{align}
\matQ_k = \vecsigma_k\trans \diag{\matQ_{cv}^{*2}, \matI_2} \vecsigma_k.
\end{align}
The key to making the model distance-aware, is here to scale $\vecsigma_k$ with the previous object width, $w_{k - 1}$, as in \cite{Aharon2022BoT-SORT:Tracking}:
\begin{align}
\vecsigma_k &= 
w_{k - 1}
\begin{pmatrix}
  \sigma_{ca},
\, \sigma_{ca},
\, \sigma_{ca},
\, \sigma_{ca},
\, \sigma_{sr},
\, \sigma_{sr}
\end{pmatrix},
\end{align}
where $\sigma_{ca}$ and $\sigma_{sr}$ are the standard deviation of the center acceleration noise and the size rate noise, respectively.

For the sensor model we assume that we make observations corrupted by white Gaussian noise as
\begin{align}
\vecz_k &= \left(\vecx_k\right)_{c_x c_y w h} + \vecw_k,\, \vecw_k \sim \mathcal{N}(\veczero, \matR_k).
\end{align}

The sensor and transition models together describe a linear Gaussian system, which is suitable for estimation using a \ac{KF}.
When initializing new tracks, we will in addition to the above employ
\begin{align}
\widehat{\vecx}_0 &=  \begin{pmatrix}
\vecz,
&
0,
&
0
\end{pmatrix}\\
\widehat{\matP}_0 &= \diag{\matR, \matP_{cr}}
\end{align}
as a prior for the center rate, where $\matR$ and $\matP_{cr}$ are the to-be-estimated measurement- and initial center rate covariances.

\subsection{Modelling detector performance}\label{sec:clutter-model}
In probabilistic tracking, the extraneous measurement density ($\lambda_{EX}$ from \cref{eq:lambda-ex}) and the detection confidence ($\lik{\vecz_j^{(c)}}$ from \cref{eq:det-track-lik}) are typically ignored or treated as constants.
However, these quantities describe critical aspects of the detector performance that should affect the tracking.
Properly modeling these parameters can allow us to quickly establish track in the simple cases, while still avoiding false tracks in questionable cases.

We begin by modeling the detector confidence $\vecz^{(c)}$ in \cref{eq:det-track-lik} through histogram binning of inlier and all detections on the training set (see \cref{sec:estimate-motion} for details) as
\begin{align}
\lik{\vecz_j^{(c)}} \triangleq \frac{\mathrm{hist}_\mathrm{inlier}(c_j, w_j)}{\mathrm{hist}_\mathrm{all}(c_j, w_j)}, \label{eq:detector-conf}
\end{align}
where $w_j$ is the measured width, and $c_j$ is the predicted confidence from the detector.
\Cref{fig:mot17-pc} shows the resulting likelihood on MOT17, which is clearly nonuniform.

Next, we shall model $\lambda_{EX}(\vecz)$, which is the density of clutter measurements and measurements due to newly appeared targets.
Through experimentation, we have found that object size alone is a good discriminator for $\lambda_{EX}$.
We therefore use scaled histogram binning over object width across \textit{all} training set detections as a pragmatic model:
\begin{align}
\lambda_{EX}(\vecz_j) \triangleq c_{EX} \mathrm{hist}_w(w_j), \label{eq:ex-hist}
\end{align}
where $w_j$ is the measured width and $c_{EX}$ is a constant we estimate in \cref{sec:estimate-motion}.
\Cref{fig:mot17-pw} shows a log-plot of $\mathrm{hist}_w$ for MOT17, which is clearly skewed towards smaller boxes.

\subsection{Automatically estimating model parameters} \label{sec:estimate-motion}
To start using the proposed motion and sensor model, we need the parameters $\sigma_{ca}$, $\sigma_{sr}$, $\matR$ and $\matP_{cr}$.
Fortunately, these can be estimated from a dataset consisting of detections and ground-truth tracks, such as those provided in the MOTChallenges.

The ground truth bounding boxes are given as $\vecg_{i}^{(k)} = (c_x, c_y, w, h)$ for each true target $x_i$ in each frame $k$ where $x_i$ is present.
We start with the prior center rate covariance $\matP_{cr}$, which can be estimated from the ground truth tracks alone.
To avoid potential errors in the camera ego-motion correction tainting the ground truth data, we only use sequences where the camera is stationary.
We then estimate $\matP_{cr}$ as
\begin{align}
\resizebox{.85\hsize}{!}{$
\matP_{cr} = \frac{1}{n_g - 1} \sum_i \frac{(\vecg_i^{(k_2)} - \vecg_i^{(k_1)})(\vecg_i^{(k_2)} - \vecg_i^{(k_1)})\trans}{(t_i^{(k_2)} - t_i^{(k_1)})^2}, $}\label{eq:pcr}
\end{align}
where $k_1$ and $k_2$ index the frames where target $i$ appears for the first and second time, $t_i^{(k_2)} - t_i^{(k_1)}$ is the time elapsed between said frames, and $n_g$ is the total number of targets.

The detection bounding boxes are given as a set $\vecz_j = (\widehat{c}_x, \widehat{c}_y, \widehat{w}, \widehat{h}) \in \mathcal{Z}$ for each frame.
To leverage the detections in parameter estimation, we first attempt to associate detections $\vecz_j$ to ground truth targets $\vecg_i$ using \ac{IOU}.
We only consider pairs where $\mathrm{iou}(\vecz_j, \vecg_i) > 0.7$, and match $\vecz_j$ to $\vecg_i$ if $\vecz_j$ is the closest to $\vecg_i$ and vice-versa.
We denote the resulting set of associations $\mathcal{A} = \left\{\left(\vecz_{j_l}, \vecg_{i_l}\right)\right\}_l$.

Using the associated detections and ground truth targets, we once again use stationary sequences, and estimate $\matR$ as
\begin{align}
\resizebox{.85\hsize}{!}{$
\matR = \frac{1}{n_a - 1} \sum_{\left(\vecz^{(k)}_j, \vecg^{(k)}_i\right) \in \mathcal{A}_k} (\vecz^{(k)}_j - \vecg^{(k)}_i)(\vecz^{(k)}_j - \vecg^{(k)}_i)\trans,
$}
\label{eq:matr}
\end{align}
where $n_a$ is the total number of associations.

The histograms over all detections as a function of box width ($\mathrm{hist}_w(w)$) and as a function of both predicted confidence and box width ($\mathrm{hist}_\mathrm{all}(c, w)$) can be computed directly from the training set.
We compute inlier histogram ($\mathrm{hist}_\mathrm{all}(c, w)$) using only the associated detections in $\mathcal{A}$.

To estimate $\sigma_{ca}$ and $\sigma_{sr}$ we employ \ac{MLE} based on $\mathcal{A}$ and the proposed motion- and sensor models from \cref{sec:motion-sensor-model}, as outlined in \cite{brekke2019}.
Finally, we find $c_{EX}$ by a parameter search where we run the tracker on the full training set.

%% file: tex/05-experiments.tex
\section{Experiments}
\label{sec:experiments}
We will now assess the effectiveness of our proposed minimalist probabilistic visual tracker.
To this end we evaluate BASE on the MOT17 \cite{Milan2016MOT16:Tracking}, MOT20 \cite{Dendorfer2020MOT20:Scenes} and DanceTrack \cite{Sun2022DanceTrack:Motion} benchmarks.
Since we are primarily interested in validating the probabilistic backbone, we opt not to use Re-Id or other appearance features.
We will focus on the MOTChallenge benchmarks, which are ideal for demonstrating a minimal probabilistic visual tracker as they contain single-camera footage with simple ego-motion and no complex movement patterns.
The DanceTrack dataset contains much more sudden movement of arms and legs, and would benefit from a more specialized motion model.

\subsection{MOTChallenge caveats}
Although the MOTChallenge benchmarks enable objective comparison of tracking algorithms, there are a few noteworthy differences in practices that color the results.

\noindent\textbf{Not all results use global parameters.}
As discussed in \cite{Cetintas2023UnifyingHierarchies}, several submissions boost performance by tuning separate parameter sets for each sequence in the \textit{test set}.
Unless explicitly stated or evident from published source code, we cannot ascertain which practice is used for a given method.

\noindent\textbf{Post-tracking interpolation} across frames where objects are not observed, is ubiquitous among all top-scoring methods on both MOT17 and MOT20.
Approaches that interpolate over a fixed number of frames, still seem to consider themselves ``online''.

\noindent\textbf{The \textit{public} detection leaderboard} seems useful to compare trackers on equal terms.
However, all the top submissions in this category still use image data to extract additional detections or Re-ID, greatly occluding the results.

\subsection{Detector}
We use the YOLOX \cite{yolox2021} detector with a detection threshold of 0.1 for all benchmarks.
For DanceTrack we use the officially trained weights.
For MOT17 and MOT20 we use weights trained in a similar fashion as for ByteTrack \cite{Zhang2021ByteTrack:Box} with a combination of MOT17, Cityperson \cite{Zhang2017CityPersons:Detection}, Crowdhuman \cite{Shao2018CrowdHuman:Crowd} and ETHZ \cite{Ess2008ATracking}. 
We excluded fully occluded targets in the training process to avoid overfitting resulting in clutter detections during large occlusions.




%


\subsection{Parameter estimation}
For each of the three benchmarks we select a global set of parameters that are used across all sequences.
Following the scheme outlined in \cref{sec:estimate-motion,sec:clutter-model} we compute $\matP_{cr}$, $\matR$, $\lik{\vecz^{(c)}}$ and histograms. 
$\sigma_{ca}$ and $\sigma_{sr}$ are estimated using MLE on inlier detections, and $c_{EX}$ is found by a parameter search using full tracking on the respective training sets.

For all benchmarks we use $P_D = 0.95$.
MOT17 and MOT20 use a canonical $P_G = 10^{-3}$.
DanceTrack has very few distracting elements, and we have seen improved performance using $P_G = 10^{-6}$.

\begin{figure}
    \centering
    \includegraphics[width=\linewidth,trim={0.4cm 0.4cm 0.4cm 0.4cm},clip]{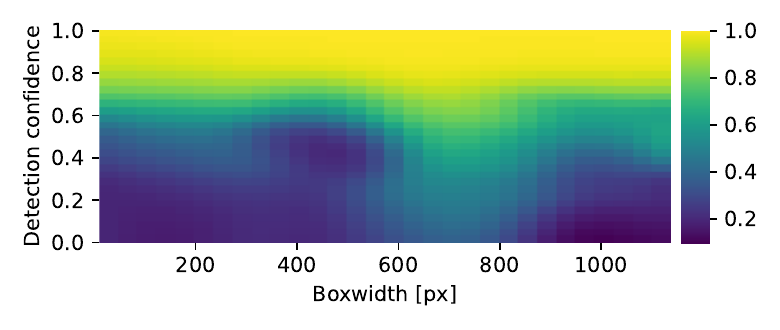}
    \caption{The $\lik{\vecz_j^{(c)}}$ histogram from \cref{eq:detector-conf} for the MOT17 training set. Each cell is the number of inlier detections (as defined in \cref{sec:estimate-motion}) divided by the number of detections within the corresponding boxwidth/score bin.}
    \label{fig:mot17-pc}
\end{figure}

\begin{figure}
    \centering
    \includegraphics[width=\linewidth,trim={0.0cm 0.2cm 0.0cm 0.2cm},clip]{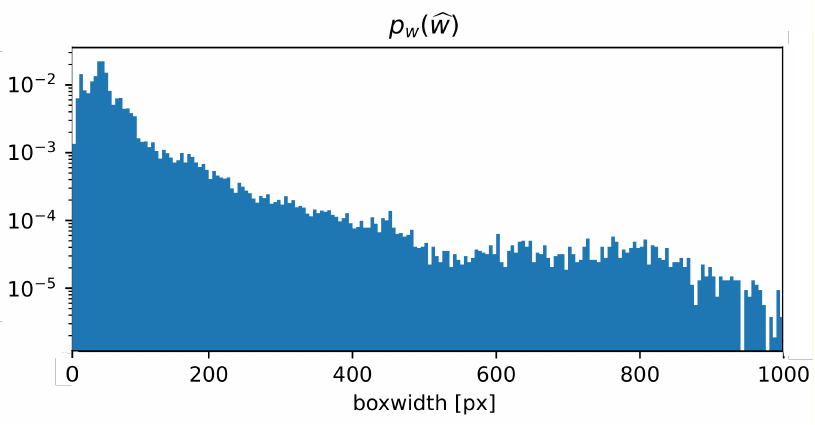}
    \caption{The $\mathrm{hist}_w(w_j)$ histogram from \cref{eq:ex-hist} for the MOT17 training set. Each cell is just the density of detections within the corresponding boxwidth bin.}
    \label{fig:mot17-pw}
\end{figure}

\subsection{Post-processing}\label{sec:post-processing}
The MOTChallenge ground truth used for scoring contains both visible and occluded targets, which makes it important to maintain tracks through occlusions. 
True real-time trackers will always have a disadvantage in this regard, as they cannot correct already reported trajectories when old targets reappear. 
To counter this, it has become a common practice among the top performing tracker to apply interpolation across such occlusion gaps in a post-processing step. With this post-processing the tracker can still run live, but will have some fixed delay.

We employ the interpolation post-processing as described in \cite{Zhang2021ByteTrack:Box}, in addition to what we call \textit{look-ahead}.
With look-ahead we delay reporting of tracks by a fixed number of frames, but use the estimated track hypothesis likelihood of the newest processed frame to determine whether the track should be reported.
The primary benefit of this is that we can report tracks with certainty upon first detection, even though the track hypothesis likelihood requires a few frames to accumulate.
Without look-ahead, the metrics used in the MOTChallenges force the tracker to establish tracks on the very first detections, severely limiting the ability of the probabilistic track management to filter clutter.


\subsection{MOT17 results}
Our overall results for MOT17 are presented in \cref{fig:mot-results,tab:mot-results}.
On the testset, \ac{BASE} (excl. the detector) ran at 331Hz on an AMD 5950x.
Among all submissions with publications and which use global parameters, \ac{BASE} ranks first with an $81.9$ MOTA score and third with a $64.5$ HOTA score.
In MOTA, \ac{BASE} is only surpassed by ImprAsso \cite{StadlerAnTracking}, which transparently report using per-sequence tuning on the test set.
We strongly suspect that \ac{BASE} would receive a significant performance boost using a similar tuning scheme, but we insist on using global parameters.
Compared to leading trackers, like BoT-SORT \cite{Aharon2022BoT-SORT:Tracking}, \ac{BASE} performs exceedingly well on sequences with small objects and persistent clutter detections, like sequence 01 ($72.4$ vs $63.4$ MOTA) and 14 ($68.1$ vs $53.5$ MOTA).
We believe this owes to \ac{BASE}'s probabilistic track management, which better captures the nuance between faint detections of small objects and more inconsistent clutter detections.

Compared to BoT-SORT, \ac{BASE} performs worse on sequences with long occlusions paired with large camera motion, like sequence 06 ($64.8$ vs $66.6$ MOTA).
We suspect this is due to our lack of Re-ID, as other approaches without Re-ID, such as ByteTrack \cite{Zhang2021ByteTrack:Box} and OC-SORT \cite{Cao2022Observation-CentricTracking}, also perform poorly on this sequence ($60.2$ and $57.3$ MOTA).

The gray section of \cref{tab:mot-results} reports results strictly using only the public detections of MOT17.
Here, \ac{BASE} outperforms vastly more complicated and computationally costly probabilistic trackers, even without \ac{VO}.

\begin{table}
{
    \footnotesize
    \SetTblrInner{rowsep=1pt,colsep=2pt}
    \centering
    \captionof{table}{\label{tab:mot-results}
    \ac{SOTA} and select methods on key benchmarks, sorted by MOT17 MOTA.
    The section with grey background show top results using strictly public detections.
    Results which tune parameters for each sequence in the test set are shown in grey font.
    }
    \begin{tblr}{
    colspec = {c|c|cc| c c | c c},
    column{3} = {colsep=1.5pt},
    column{4} = {colsep=1.5pt},
    column{5} = {colsep=1.5pt},
    column{6} = {colsep=1.5pt},
    column{7} = {colsep=1.5pt},
    column{8} = {colsep=1.5pt},
    row{3} = {bg=black!20},
    row{4} = {bg=black!20},
    row{5} = {bg=black!20},
    row{6} = {bg=black!20},
    row{11} = {fg=black!40},
    row{12} = {fg=black!40},
    row{16} = {fg=black!40},
    cell{1}{3,5,7} = {c=2}{c}
    }
    \hline
    &
    & \textbf{MOT17} &
    & \textbf{MOT20} &
    & \textbf{Dancetrack} &
    \\
    \textbf{Method}
    & \textbf{ReID}
    & \textbf{MOTA}
    & \textbf{HOTA}
    & \textbf{MOTA}
    & \textbf{HOTA}
    & \textbf{MOTA}
    & \textbf{HOTA}
    \\
    \hline     
    PHD\_GM\cite{Sanchez-Matilla2020MotionTracking}
    & - 
    & 48.8 & - & - & - & - & -
    \\
    GMPHDO.\cite{Song2019OnlineManagement}
    & - 
    & 49.9 & - & - & - & - & -
    \\
    \textbf{BASE} (no VO)
    & - 
    & 50.2 & 41.2 & - & - & - & -
    \\
    \textbf{BASE} (ours)
    & - 
    & 51.8 & 43.6 & - & - & - & -
    \\
    \hline     
    OCSORT\cite{Cao2022Observation-CentricTracking} 
         & -
    & 78.0          & 63.2    
    & 75.7          & 62.4    
    & 92.0          & 55.7    
    \\
    MOTRv2\cite{Zhang2023MOTRv2:Detectors}
        & \cmark
    & 78.6          & 62.0     
    & 76.2          & 60.3           
    & 92.1          & 73.4     
    \\
    StrongSort\cite{Du2022StrongSORT:Again} 
        & \cmark
    & 79.6          & 64.4         
    & 73.8          & 62.6         
    & -              &-            
    \\
    FineTrack\cite{RenFocusRepresentation}
        & \cmark
    & 80.0          & 64.3
    & 77.9          & 63.6
    & 89.9              &52.7
    \\
    ByteTrack\graycite{Zhang2021ByteTrack:Box} 
         & -
    & 80.3          & 63.1    
    & 77.8          & 61.3    
    & 90.9          & 51.9      
    \\
    BoT-SORT\graycite{Aharon2022BoT-SORT:Tracking}
        & \cmark
    & 80.5          & 65.0 
    & 77.8          & 63.3 
        & -              &-            
    \\
    SUSHI\cite{Cetintas2023UnifyingHierarchies}
        & \cmark
    & 81.1          & 66.5
    & 74.3          & 64.3
    & 88.7          & 63.3
    \\
    MotionT.\cite{QinMotionTrack:Tracking} 
        & \cmark
    & 81.1          & 65.1
    & 78.0          & 62.8
    & -             & -
    \\
    CBIoU\cite{Yang2023HardSpace}
        & \cmark
    & 81.1          & 64.1 
        & -              &-               
    & 91.6          & 60.6 
    \\
    ImprAsso\graycite{StadlerAnTracking}
        & \cmark
    & 82.2          & 66.4
    & 78.6          & 64.6
    & -              &-
    \\
    \hline
    \textbf{BASE} (ours)
    & -
    & 81.9 & 64.5 
    & 78.2   & 63.5   
        & 91.7  & 56.4 
    \\
    \hline
    \end{tblr}
    
\hspace{0.05\textwidth}
}
\end{table}

\subsection{MOT20 results}
The results for MOT20 are also shown in \cref{fig:mot-results,tab:mot-results}.
Among submissions with publications and which use global parameters, \ac{BASE} ranks first in MOTA, and third in HOTA.
This is somewhat surprising, given that several of the other top methods employ Re-Id, which seems particularly promising on MOT20.
The high HOTA-score might indicate that an empirically tuned motion model helps prevent mixing up targets during occlusions.
On the testset, \ac{BASE} (excl. the detector) ran at 39Hz on an AMD 5950x.

Several of the top-performing trackers employ per-sequence parameters, which seems to be particularly beneficial on MOT20.
Sequences 04 and 07, the scenes of which are also featured in the training set, seem to warrant conservative tracker parameters.
Meanwhile, sequences 06 and 08 seem to benefit from a more sensitive tracker.

\subsection{DanceTrack}
DanceTrack \cite{Sun2022DanceTrack:Motion} is an interesting dataset as it poses quite different challenges than MOT17 and MOT20, with highly irregular motion but often relatively easily detectable targets.
Since \ac{BASE}'s motion model assumes continuous motion and slow changes, the sudden streching of arms and changes in posture typical for this dataset seems particularly ill-suited for our model.
\Cref{tab:mot-results} shows our results, as well as the \ac{SOTA} methods also using the public detector.
Our method outperforms ByteTrack\cite{Zhang2021ByteTrack:Box} and FineTrack\cite{RenFocusRepresentation} but falls behind C-BIoU, which leverages a more specialized motion model. 
OCSORT\cite{Cao2022Observation-CentricTracking}, which also uses a motion model adapted to DanceTrack, performs similarly to BASE. 
MOTRv2\cite{Zhang2023MOTRv2:Detectors} outperforms all these approaches by a large margin, illustrating that an end-to-end approach may be a particularly good fit for DanceTrack.
However, as they use additional training data, the MOTRv2 results are not directly comparable to the other methods.

\subsection{Ablation study}
In this section, we study the effects of the key components of \ac{BASE}, namely the probabilistic association (vs IOU-based assocation), the distance-aware motion model (vs naive motion model), dynamic clutter estimation (vs constant $\lambda_\mathrm{EX}$) and the histogram-calibrated detector confidence (vs ignoring or using raw confidence).
We use the same YOLOX ablation model from \cite{Zhang2021ByteTrack:Box}, so the ablation results are directly comparable to those of BoT-SORT \cite{Aharon2022BoT-SORT:Tracking} and ByteTrack \cite{Zhang2021ByteTrack:Box}.
The model was trained on Crowdhuman \cite{Shao2018CrowdHuman:Crowd} and the \verb#train# half of the MOT17 training set.
We fit all \ac{BASE}-specific parameters using only \verb#train# from MOT17, while the experiments were run on the \verb#val# half of the training set.
The results are shown in \cref{tab:ablation-results}.

Comparing configurations 1 and 2, we see that when using a constant $\lambda_{EX}$ instead of dynamic clutter (\cref{eq:ex-hist}), the distance-aware motion model actually makes the method perform worse.
Meanwhile, comparing row 3 to row 6, we see that the distance-aware motion model gives a significant boost once the dynamic clutter model is in place.
Since the distance-aware motion model increases the position uncertainty for tracks with large bounding boxes, such tracks struggle to build confidence and match with detections when the clutter density is kept constant.

We also see that probabilistic association with a naive motion model (row 3) is outperformed by IOU-based association (row 4).
Since the IOU-based association intrinsically offers some compensation for distance (larger boxes are allowed to miss by more pixels while still achieving the same IOU as smaller boxes), this result indicates that using some distance-aware mechanism is indeed necessary in visual tracking.
Meanwhile, we also see that our proposed distance-aware motion model with probabilistic association (row 6) performs even better, indicating that \ac{BASE} is able to further exploit the distance information.

We observe a significant improvement in performance when calibrated confidence is used (row 8) compared to raw confidence (row 7).
Ignoring the confidence score altogether (row 6) results in worse performance than BoT-SORT.
A possible explanation for this is that the BoT-SORT dual threshold approach is able to extract some, but not all, of the confidence score information.
Since all proposed components are necessary to reach \ac{SOTA} in the probabilistic paradigm, we consider \ac{BASE} as a minimalist approach.

\begin{table}
\setlength{\tabcolsep}{4.3pt}
\caption{\label{tab:ablation-results}
Ablation results on the MOT17 validation set.
The experiments quantify the effects of our proposed dynamic clutter density model, the distance-aware motion model, the probabilistic association and the use of detector confidence.
}
\footnotesize
\begin{tabular}{r|cccc|cc}
\hline

& \textbf{\makecell{Dynamic\\ clutter}}
& \textbf{\makecell{Distance\\-aware}}
& \textbf{\makecell{Prob.\\assoc.}}
& \textbf{\makecell{Detection\\confidence}}
& \textbf{\makecell{\\MOTA}}
& \textbf{\makecell{\\HOTA}}
\\ \hline
\textbf{1} & -          & \checkmark & \checkmark    & -     & 69.3 & 65.7 \\
\textbf{2} & -          & -          & \checkmark    & -     & 71.7 & 65.7 \\
\textbf{3} & \checkmark & -          & \checkmark    & -     & 75.6 & 66.4 \\
\textbf{4} & \checkmark & -          & IOU           & -     & 76.0 & 66.1 \\
\textbf{5} & \checkmark & \checkmark & IOU           & -     & 76.5 & 66.2 \\
\textbf{6} & \checkmark & \checkmark & \checkmark    & -     & 77.1 & 68.2 \\
\rowcolor{lightgray} & \multicolumn{4}{c|}{BoT-SORT\cite{Aharon2022BoT-SORT:Tracking}} & 78.5 & 69.2 \\ 
\textbf{7} & \checkmark & \checkmark & \checkmark    & Raw   & 80.8 & 69.8 \\
\textbf{8} & \checkmark & \checkmark & \checkmark    & Calib & 81.6 & 70.2 \\
\end{tabular}
\end{table}

%% file: tex/06-conclusion.tex
\section{Conclusion}
\label{sec:conclusione}
This paper demonstrates that a probabilistic tracker can achieve \ac{SOTA} on popular \ac{VMOT} benchmarks.
Our proposed method, \ac{BASE}, is the top-performing method on the MOT17 and MOT20 benchmarks and has competitive results on the more specialized DanceTrack benchmark.
Starting from a minimalist probabilistic \ac{SHT} foundation, we found that the components identified in \cref{sec:introduction} significantly contribute to the overall performance.
Our ablation study shows that the probabilistic association relies on using a distance-aware motion model to perform well, which in turn relies on proper (dynamic) clutter modeling to function.
Including calibrated detector confidence scores further improves the results beyond the current SOTA.

Probabilistic trackers are particularly well suited for exploiting multiple object features, such as appearance/Re-ID.
Our goal is that \ac{BASE} can serve as starting point for more advanced probabilistic visual trackers.

%% file: main.bbl
\begin{thebibliography}{10}\itemsep=-1pt

\bibitem{Aguilar2022SmallFilter}
Camilo Aguilar, Mathias Ortner, and Josiane Zerubia.
\newblock {Small Object Detection and Tracking in Satellite Videos With Motion
  Informed-CNN and GM-PHD Filter}.
\newblock {\em Frontiers in Signal Processing}, 2(April):1--15, 2022.

\bibitem{Aharon2022BoT-SORT:Tracking}
Nir Aharon, Roy Orfaig, and Ben-Zion Bobrovsky.
\newblock {BoT-SORT: Robust Associations Multi-Pedestrian Tracking}.
\newblock {\em arXiv preprint arXiv:2206.14651}, 6 2022.

\bibitem{Baisa2019OnlineLearning}
Nathanael~L. Baisa.
\newblock {Online Multi-object Visual Tracking using a GM-PHD Filter with Deep
  Appearance Learning}.
\newblock {\em FUSION 2019 - 22nd International Conference on Information
  Fusion}, 2019.

\bibitem{Baisa2021Occlusion-robustRe-identification}
Nathanael~L. Baisa.
\newblock {Occlusion-robust online multi-object visual tracking using a GM-PHD
  filter with CNN-based re-identification}.
\newblock {\em Journal of Visual Communication and Image Representation},
  80(January):103279, 2021.

\bibitem{Baisa2021RobustLearning}
Nathanael~L. Baisa.
\newblock {Robust online multi-target visual tracking using a HISP filter with
  discriminative deep appearance learning}.
\newblock {\em Journal of Visual Communication and Image Representation}, 77,
  2021.

\bibitem{Bar-Shalom2007DimensionlessTracking}
Yaakov Bar-Shalom, Sam~S. Blackman, and Robert~J. Fitzgerald.
\newblock {Dimensionless score function for multiple hypothesis tracking}.
\newblock {\em IEEE Transactions on Aerospace and Electronic Systems},
  43(1):392--400, 2007.

\bibitem{Bar-Shalom1975TrackingAssociation}
Yaakov Bar-Shalom and Edison Tse.
\newblock {Tracking in a cluttered environment with probabilistic data
  association}.
\newblock {\em Automatica}, 11(5), 1975.

\bibitem{Bewley2016SimpleTracking}
Alex Bewley, Zongyuan Ge, Lionel Ott, Fabio Ramos, and Ben Upcroft.
\newblock {Simple Online and Realtime Tracking}.
\newblock {\em IEEE international conference on image processing (ICIP)}, 2
  2016.

\bibitem{blackman1999design}
Samuel~S Blackman and Robert Popoli.
\newblock {\em {Design and analysis of modern tracking systems}}, volume 1999.
\newblock Artech House Publishers, 1999.

\bibitem{Bochinski2017High-SpeedInformation}
Erik Bochinski, Volker Eiselein, and Thomas Sikora.
\newblock {High-Speed tracking-by-detection without using image information}.
\newblock In {\em 2017 14th IEEE International Conference on Advanced Video and
  Signal Based Surveillance, AVSS 2017}, 2017.

\bibitem{brekke2019}
E.~F. Brekke.
\newblock {\em {Fundamentals of Sensor Fusion: Target tracking, Navigation and
  SLAM}}.
\newblock NTNU, 2019.

\bibitem{Cao2022Observation-CentricTracking}
Jinkun Cao, Xinshuo Weng, Rawal Khirodkar, Jiangmiao Pang, and Kris Kitani.
\newblock {Observation-Centric SORT: Rethinking SORT for Robust Multi-Object
  Tracking}.
\newblock {\em arXiv preprint arXiv:2203.14360}, 2022.

\bibitem{Cetintas2023UnifyingHierarchies}
Orcun Cetintas, Guillem Bras{\'{o}}, and Laura Leal-Taix{\'{e}}.
\newblock {Unifying Short and Long-Term Tracking With Graph Hierarchies}, 2023.

\bibitem{Dendorfer2020MOT20:Scenes}
Patrick Dendorfer, Hamid Rezatofighi, Anton Milan, Javen Shi, Daniel Cremers,
  Ian Reid, Stefan Roth, Konrad Schindler, and Laura Leal-Taix{\'{e}}.
\newblock {MOT20: A benchmark for multi object tracking in crowded scenes}.
\newblock {\em arXiv preprint arXiv:2003.09003}, 3 2020.

\bibitem{Du2022StrongSORT:Again}
Yunhao Du, Yang Song, Bo Yang, and Yanyun Zhao.
\newblock {StrongSORT: Make DeepSORT Great Again}.
\newblock {\em arXiv}, 2 2022.

\bibitem{Ess2008ATracking}
Andreas Ess, Bastian Leibe, Konrad Schindler, and Luc Van~Gool.
\newblock {A mobile vision system for robust multi-person tracking}.
\newblock {\em 26th IEEE Conference on Computer Vision and Pattern Recognition,
  CVPR}, 2008.

\bibitem{Fu2019Multi-LevelTracking}
Zeyu Fu, Federico Angelini, Jonathon Chambers, and Syed~Mohsen Naqvi.
\newblock {Multi-Level Cooperative Fusion of GM-PHD Filters for Online Multiple
  Human Tracking}.
\newblock {\em IEEE Transactions on Multimedia}, 21(9):2277--2291, 2019.

\bibitem{Fu2018ParticleLearning}
Zeyu Fu, Pengming Feng, Federico Angelini, Jonathon Chambers, and Syed~Mohsen
  Naqvi.
\newblock {Particle PHD Filter Based Multiple Human Tracking Using Online
  Group-Structured Dictionary Learning}.
\newblock {\em IEEE Access}, 6, 2018.

\bibitem{yolox2021}
Zheng Ge, Songtao Liu, Feng Wang, Zeming Li, and Jian Sun.
\newblock {YOLOX: Exceeding YOLO Series in 2021}.
\newblock {\em arXiv}, 7 2021.

\bibitem{JinlongYang2022ImprovingFilters}
{Jinlong Yang}, {Peng Ni}, {Jiani Miao}, and {Hongwei Ge}.
\newblock {Improving visual multi‐object tracking algorithm via integrating
  GM‐PHD and correlation filters}, 2022.

\bibitem{Kim2015MultipleRevisited}
Chanho Kim, Fuxin Li, Arridhana Ciptadi, and James~M. Rehg.
\newblock {Multiple Hypothesis Tracking Revisited}.
\newblock In {\em 2015 IEEE International Conference on Computer Vision
  (ICCV)}, pages 4696--4704. IEEE, 12 2015.

\bibitem{Leal-Taixe2015MOTChallengeTracking}
Laura Leal-Taix{\'{e}}, Anton Milan, Ian Reid, Stefan Roth, and Konrad
  Schindler.
\newblock {MOTChallenge 2015: Towards a Benchmark for Multi-Target Tracking}.
\newblock {\em arXiv}, 4 2015.

\bibitem{Mahler2003MultitargetMoments}
Ronald P~S Mahler and Lockheed Martin.
\newblock {Multitarget Bayes Filtering via First-Order Multitarget Moments}.
\newblock {\em IEEE TRANSACTIONS ON AEROSPACE AND ELECTRONIC SYSTEMS}, 39(4),
  2003.

\bibitem{Milan2016MOT16:Tracking}
Anton Milan, Laura Leal-Taixe, Ian Reid, Stefan Roth, and Konrad Schindler.
\newblock {MOT16: A Benchmark for Multi-Object Tracking}.
\newblock {\em arXiv}, 3 2016.

\bibitem{Nasseri2022FastTracking}
Mohammad~Hossein Nasseri, Mohammadreza Babaee, Hadi Moradi, and Reshad
  Hosseini.
\newblock {Fast Online and Relational Tracking}.
\newblock {\em arXiv}, 2022.

\bibitem{QinMotionTrack:Tracking}
Zheng Qin, Sanping Zhou, Le Wang, Jinghai Duan, Gang Hua, and Wei Tang.
\newblock {MotionTrack: Learning Robust Short-term and Long-term Motions for
  Multi-Object Tracking}.
\newblock In {\em Proceedings of the IEEE/CVF Conference on Computer Vision and
  Pattern Recognition}, pages 17939--17948, 3 2023.

\bibitem{Reid1979AnTargets}
Donald~B. Reid.
\newblock {An Algorithm for Tracking Multiple Targets}.
\newblock {\em IEEE Transactions on Automatic Control}, 24(6):843--854, 1979.

\bibitem{RenFocusRepresentation}
Hao Ren, Shoudong Han, Huilin Ding, Ziwen Zhang, Hongwei Wang, and Faquan Wang.
\newblock {Focus On Details: Online Multi-object Tracking with Diverse
  Fine-grained Representation}.
\newblock In {\em Proceedings of the IEEE/CVF Conference on Computer Vision and
  Pattern Recognition}, pages 11289--11298, 2 2023.

\bibitem{Sanchez-Matilla2020MotionTracking}
Ricardo Sanchez-Matilla and Andrea Cavallaro.
\newblock {Motion Prediction for First-Person Vision Multi-object Tracking}.
\newblock In {\em Lecture Notes in Computer Science (including subseries
  Lecture Notes in Artificial Intelligence and Lecture Notes in
  Bioinformatics)}, volume 12538 LNCS, pages 485--499. Springer Science and
  Business Media Deutschland GmbH, 2020.

\bibitem{Sanchez-Matilla2016OnlineDetections}
Ricardo Sanchez-Matilla, Fabio Poiesi, and Andrea Cavallaro.
\newblock {Online multi-target tracking with strong and weak detections}.
\newblock {\em Lecture Notes in Computer Science (including subseries Lecture
  Notes in Artificial Intelligence and Lecture Notes in Bioinformatics)}, 9914
  LNCS:84--99, 2016.

\bibitem{Shao2018CrowdHuman:Crowd}
Shuai Shao, Zijian Zhao, Boxun Li, Tete Xiao, Gang Yu, Xiangyu Zhang, and Jian
  Sun.
\newblock {CrowdHuman: A Benchmark for Detecting Human in a Crowd}.
\newblock {\em arXiv}, pages 1--9, 2018.

\bibitem{Song2019OnlineManagement}
Young-Min Song, Kwangjin Yoon, Young-Chul Yoon, Kin~Choong Yow, and Moongu
  Jeon.
\newblock {Online Multi-Object Tracking With GMPHD Filter and Occlusion Group
  Management}.
\newblock {\em IEEE Access}, 7:165103--165121, 2019.

\bibitem{Stadler2022ModellingCrowds}
Daniel Stadler and Jurgen Beyerer.
\newblock {Modelling Ambiguous Assignments for Multi-Person Tracking in
  Crowds}.
\newblock {\em Proceedings - 2022 IEEE/CVF Winter Conference on Applications of
  Computer Vision Workshops, WACVW 2022}, pages 133--142, 2022.

\bibitem{StadlerAnTracking}
Daniel Stadler and Jürgen Beyerer.
\newblock {An Improved Association Pipeline for Multi-Person Tracking}.
\newblock In {\em 2023 IEEE/CVF Conference on Computer Vision and Pattern
  Recognition Workshops (CVPRW)}, pages 3170--3179. IEEE, 6 2023.

\bibitem{Sun2022DanceTrack:Motion}
Peize Sun, Jinkun Cao, Yi Jiang, Zehuan Yuan, Song Bai, Kris Kitani, and Ping
  Luo.
\newblock {DanceTrack: Multi-Object Tracking in Uniform Appearance and Diverse
  Motion}, 2022.

\bibitem{Sun2020TransTrack:Transformer}
Peize Sun, Jinkun Cao, Yi Jiang, Rufeng Zhang, Enze Xie, Zehuan Yuan, Changhu
  Wang, and Ping Luo.
\newblock {TransTrack: Multiple Object Tracking with Transformer}.
\newblock {\em arXiv}, 2020.

\bibitem{Wang2020JointNetworks}
Yongxin Wang, Kris Kitani, and Xinshuo Weng.
\newblock {Joint Object Detection and Multi-Object Tracking with Graph Neural
  Networks}.
\newblock {\em arXiv}, 6 2020.

\bibitem{Wojke2018SimpleMetric}
Nicolai Wojke, Alex Bewley, and Dietrich Paulus.
\newblock {Simple online and realtime tracking with a deep association metric}.
\newblock {\em Proceedings - International Conference on Image Processing,
  ICIP}, 2017-Septe:3645--3649, 2018.

\bibitem{Wojke2017Confidence-AwareTracking}
Nicolai Wojke and Dietrich Paulus.
\newblock {Confidence-Aware probability hypothesis density filter for visual
  multi-object tracking}.
\newblock {\em VISIGRAPP 2017 - Proceedings of the 12th International Joint
  Conference on Computer Vision, Imaging and Computer Graphics Theory and
  Applications}, 6(Visigrapp):132--139, 2017.

\bibitem{Yan2022TowardsTracking}
Bin Yan, Yi Jiang, Peize Sun, Dong Wang, Zehuan Yuan, Ping Luo, and Huchuan Lu.
\newblock {Towards Grand Unification of Object Tracking}.
\newblock {\em arXiv}, 7 2022.

\bibitem{Yang2021ReMOT:Tracking}
Fan Yang, Xin Chang, Sakriani Sakti, Yang Wu, and Satoshi Nakamura.
\newblock {ReMOT: A model-agnostic refinement for multiple object tracking}.
\newblock {\em Image and Vision Computing}, 106:104091, 2021.

\bibitem{Yang2023HardSpace}
Fan Yang, Shigeyuki Odashima, Shoichi Masui, and Shan Jiang Fujitsu~Research.
\newblock {Hard To Track Objects With Irregular Motions and Similar
  Appearances? Make It Easier by Buffering the Matching Space}.
\newblock In {\em Proceedings of the IEEE/CVF Winter Conference on Applications
  of Computer Vision (WACV),}, pages 4799--4808, 2023.

\bibitem{Zeng}
Fangao Zeng, Bin Dong, Yuang Zhang, Tiancai Wang, Xiangyu Zhang, and Yichen
  Wei.
\newblock {MOTR: End-to-End Multiple-Object Tracking with Transformer}.
\newblock {\em arXiv}, 5 2021.

\bibitem{Zhang2017CityPersons:Detection}
Shanshan Zhang, Rodrigo Benenson, and Bernt Schiele.
\newblock {CityPersons: A diverse dataset for pedestrian detection}.
\newblock {\em Proceedings - 30th IEEE Conference on Computer Vision and
  Pattern Recognition, CVPR 2017}, 2017-Janua:4457--4465, 2017.

\bibitem{Zhang2021ByteTrack:Box}
Yifu Zhang, Peize Sun, Yi Jiang, Dongdong Yu, Zehuan Yuan, Ping Luo, Wenyu Liu,
  and Xinggang Wang.
\newblock {ByteTrack: Multi-Object Tracking by Associating Every Detection
  Box}.
\newblock {\em ECCV 2022, Proceedings}, 10 2021.

\bibitem{Zhang2023MOTRv2:Detectors}
Yuang Zhang, Tiancai Wang, and Xiangyu Zhang.
\newblock {MOTRv2: Bootstrapping End-to-End Multi-Object Tracking by Pretrained
  Object Detectors}.
\newblock In {\em Proceedings of the IEEE/CVF Conference on Computer Vision and
  Pattern Recognition}, pages 22056--22065, 11 2023.

\bibitem{Zhu2021LookingTransformers}
Tianyu Zhu, Markus Hiller, Mahsa Ehsanpour, Rongkai Ma, Tom Drummond, Ian Reid,
  and Hamid Rezatofighi.
\newblock {Looking Beyond Two Frames: End-to-End Multi-Object Tracking Using
  Spatial and Temporal Transformers}.
\newblock {\em arXiv}, pages 1--20, 2021.

\end{thebibliography}
